\documentclass[sigconf,authorversion,nonacm]{acmart}
\usepackage{subcaption}
\usepackage{float}
\graphicspath{{Figures/}}
\usepackage{graphicx}
\usepackage[ruled,vlined,linesnumbered]{algorithm2e}

\AtBeginDocument{%
  \providecommand\BibTeX{{%
    \normalfont B\kern-0.5em{\scshape i\kern-0.25em b}\kern-0.8em\TeX}}}

\setcopyright{acmcopyright}
\copyrightyear{2018}
\acmYear{2018}
\acmDOI{10.1145/1122445.1122456}

\begin{document}

\title{Using Pareto Simulated Annealing to Address Algorithmic Bias in Machine Learning}

\author{William Blanzeisky, P\'{a}draig Cunningham}
\email{william.blanzeisky@ucdconnect.ie,  padraig.cunningham@ucd.ie}
\affiliation{%
  \institution{School of Computer Science\\
	University College Dublin}
  \city{Dublin 4}
  \country{Ireland}
}


\begin{abstract}
Algorithmic Bias can be due to bias in the training data or issues with the algorithm itself. These algorithmic issues typically relate to problems with model capacity and regularisation. This \emph{underestimation} bias may arise because the model has been optimized for good generalisation accuracy without any explicit consideration of bias or fairness. In a sense, we should not be surprised that a model might be biased when it hasn't been `asked' not to be. In this paper, we consider including bias (underestimation) as an additional criterion in model training. We present a multi-objective optimization strategy using Pareto Simulated Annealing that optimizes for both balanced accuracy \emph{and} underestimation. We demonstrate the effectiveness of this strategy on one synthetic and two real-world datasets.
\end{abstract}



\maketitle

\section{Introduction}\label{intro}

While Machine learning (ML) systems have revolutionized many aspects of human lives, the growing evidence of algorithmic bias necessitates the need for fairness-aware ML. ML algorithms rely on the training data to make predictions that often have high societal impacts, such as determining the likelihood of convicted criminals re-offending \cite{julia_angwin_2016}. Thus, algorithms that are trained on a biased representation of the actual population could disproportionately disadvantage a specific group or groups. Although most examples of algorithmic bias in ML occur due to the training data, recent work shows that the algorithm itself can introduce bias or amplify existing bias \cite{blanzeisky2021algorithmic,cunningham2020algorithmic,hooker2020characterising}. When the bias occurs due to the data, it is sometimes euphemistically called negative legacy; it is called underestimation when it is due to the algorithm \cite{cunningham2020algorithmic}. 

Over the past few years, several approaches to mitigate bias in ML have been proposed \cite{caton2020fairness}. One strategy is to modify an algorithm's objective function to account for one or more fairness measures. For example, one can enforce a fairness measure as a constraint directly into an algorithm optimization function \cite{pmlr-v28-zemel13}. From this perspective, algorithmic bias can be formulated as a multi-objective optimization problem (MOOP), where the objective is usually to maintain good predictive accuracy while ensuring fair outcomes across protected groups. The main challenge with these approaches is that they usually result in a non-convex optimization function \cite{Goel2018AAAI}. Although convexity is often required for algorithmic convenience, there are several possible approaches to address this issue; \cite{zafar2015} use the co-variance between sensitive attribute and target feature as a proxy for a convex approximate measure of fairness, \cite{zafarDDCP} convert non-convex fairness constraint into a Disciplined Convex-Concave Program (DCCP) and leverage recent advances in convex-concave programming to solve it, while \cite{DBLP:journals/corr/CotterFGG16} utilizes majorization-minimization procedure for (approximately) optimizing non-convex function.

In this paper, we focus on mitigating underestimation bias. Specifically, we formulate underestimation as a MOOP and propose a remediation strategy using Pareto Simulated Annealing (PSA) to simultaneously optimize the algorithm on two objectives: (1) maximizing prediction balance accuracy; (2) ensuring little to no underestimation of desirable outcomes (w.r.t minority group). Since the optimization in MOOP is usually a compromise between multiple competing solutions (in our case, balance accuracy and underestimation), the main objective is to use Pareto optimality to find a set of optimal solutions created by the two competing criteria \cite{mas1995microeconomic}.

The specifics of the proposed remediation strategy are presented in more detail in section \ref{sec:mitigating_underestimation}. Before that, the relevant background on key concepts used in this paper is reviewed in section \ref{sec:background:bias} and \ref{sec:background:MOOP_PSA}. The paper concludes in section \ref{sec:experiments} with an assessment of how this repair strategy works on one synthetic and two real datasets.

\section{Bias in Machine Learning}\label{sec:background:bias}
Existing research in ML bias can generally be categorized into two categories: bias discovery and bias mitigation \cite{zliobaite2017fairnessaware}. Most literature in bias discovery focuses on quantifying bias and developing theoretical understanding of the social and legal aspects of ML bias, while bias prevention focuses on technical approaches to mitigate biases in ML systems. Several notions to quantify fairness have been proposed \cite{caton2020fairness}. One of the accepted measure of unfairness is Disparate Impact (DIs) \cite{feldman2015certifying}:
\begin{equation}\label{eqn:DI}
   \mathrm{DI}_S \leftarrow \frac{P[\hat{Y}= 1 | S \ne 1]}{P[\hat{Y} = 1 \vert S = 1]} < \tau 
\end{equation}
DI$_S$ is defined as the ratio of desirable outcomes $\hat{Y}$ predicted for the sensitive minority $S \ne 1$ compared with that for the majority $S = 1$. $\tau = 0.8$ is the 80\% rule, i.e. proportion of desirable outcomes for the minority should be within 80\% of those for the majority. However, this measure emphasizes fairness for all subgroups without taking into account the source of the bias. 

As stated in section \ref{intro}, it is worth emphasizing the difference between negative legacy and underestimation as sources of bias in ML. Negative legacy refers to the problem with the data while underestimation refers to the bias due to the algorithm. Negative legacy may be due to labelling errors or poor sampling; however, it is likely to reflect discriminatory practices in the past. On the other hand, underestimation occurs when the algorithm focuses on strong signals in the data thereby missing more subtle phenomena \cite{cunningham2020algorithmic}. Recent work shows that underestimation occurs when an algorithm underfits the training data due to a combination of limitations in training data and model capacity issues \cite{cunningham2020algorithmic,blanzeisky2021algorithmic,kamishima2012fairness}. \cite{blanzeisky2021algorithmic} also shows that irreducible error, regularization and feature and class imbalance can contribute to this underestimation. 

We define \emph{underestimation score} ($\mathrm{US}_{S}$) in line with $\mathrm{DI}_S$:
\begin{equation}\label{eqn:US_S1}
    \mathrm{US}_{S} \leftarrow \frac{P[\hat{Y}= 1 | S \ne 1]}{P[Y = 1 | S \ne 1]} 
\end{equation}
This is the ratio of desirable outcomes predicted by the classifier for the sensitive minority compared with what is actually present in the data \cite{cunningham2020algorithmic}. If $\mathrm{US}_{S} <1$ the classifier is under-predicting  desirable outcomes for the minority. It is worth nothing that $\mathrm{US}_{S} = 1$ does not necessarily mean that the classifier is not biased against the minority group (i.e. poor $\mathrm{DI}_S$) score.

\section{Multi-Criteria Optimization} 
\label{sec:background:MOOP_PSA}

Multi-objective optimization problem (MOOP), or multi-criteria optimization, refers to problems where two or more objective functions have to be simultaneously optimized. Given two or more objective functions $f_{i}$ for $i = 1,2,...,m$, MOOP can generally be formulated as:
\begin{equation}
    \min_x (f_{1}(x),f_{2}(x),f_{3}(x),...,f_{m}(x)) 
\end{equation}
As stated in Section \ref{intro}, ensuring fairness in ML algorithms can be formulated as a MOOP. To solve this, we have to consider the fact that the two objectives might be competing in such a way that no improvement on one objective is possible without making it worse in any one of the others. Thus, if there is no single solution that dominates in both criteria , MOOP uses the concept of Pareto optimality to find a set of non-dominated solutions created by the two competing criteria  \cite{mas1995microeconomic}. This set of solution is also referred to as the Pareto set.

\subsection{Simulated Annealing}
Simulated annealing (SA) is a single-objective meta-heuristic to approximate a global optimization function with a very large search space \cite{Kirkpatrick671}. SA is similar to stochastic hill-climbing but with provision to allow for worse solutions to be accepted with some probability \cite{foley}. Inspired by the natural process of annealing solids in metallurgy, the acceptance of inferiors solutions is controlled by a temperature variable $T$ so that it will becomes increasingly unlikely as the system cools down. This will allow SA to jump out of local minima, which is often desirable when optimizing non-convex functions.

Given an objective function $f(x)$, initial solution $x$, and initial temperature $T$, the simulated annealing process consists of first finding a neighbor $\bar{x}$ as a candidate solution by perturbing the initial solution $x$. If the candidate solution $\bar{x}$ improves on $x$, then it is accepted with a probability of 1. In contrast, if $\bar{x}$ is worse than $x$, it may still be accepted with a probability $P(x,\bar{x},T)$. The acceptance probability for inferior solutions follows a Boltzmann probability distribution and can be defined as \cite{https://doi.org/10.1002/(SICI)1099-1360(199801)7:1<34::AID-MCDA161>3.0.CO;2-6}:
\begin{equation}
    P(x,\bar{x},T) = \min \left(1,exp\left(\frac{f(x) - f(\bar{x})}{T}\right) \right)
\end{equation}

The evaluation process is repeated $n_{iter}$ times for each temperature T, while the decrease in T is controlled by a cooling rate $\alpha$. The process stops once changes stop being accepted.

\subsection{Pareto Simulated Annealing}

Pareto Simulated Annealing (PSA) is an extension of SA for handling MOOP by exploiting the idea of constructing an estimated Pareto set \cite{article}. Instead of starting with one solution, PSA initializes a set of solutions. Candidate solutions are generated from this set to obtain a diversified Pareto front. PSA aggregates the acceptance probability for all the competing criteria so that the acceptance probability of inferior solutions is defined as \cite{https://doi.org/10.1002/(SICI)1099-1360(199801)7:1<34::AID-MCDA161>3.0.CO;2-6}:

\begin{equation}\label{eqn:prob_accept}
    P(x,\bar{x},T,\lambda) = \acmISBN{}min \left\{ 1,exp\left(\frac{\sum_{i=0}^{m} \lambda_{i} (f_{i}(x) - f_{i}(\bar{x}))}{T}\right) \right\}
\end{equation}

where $m$ is the number of objectives to be optimized and $\lambda$ is a weight vector that represents the varying magnitude of importance of the objectives. Depending on the units of the criteria, it may be desirable to normalize the function so that the movement of a function in one solution when compared with the function in the original solutions is treated as percentage improvements \cite{foley}. In other words, the differences in particular objectives for acceptance criteria probability calculation are aggregated with a simple weighted sum. It is worth noting that the temperature $T$ is dependent on the range of the objective functions.

PSA has shown significant success for many applications; designing optimal water distribution networks \cite{https://doi.org/10.1029/2019WR025852}, optimizing control cabinet layout \cite{8745108}, optimizing for accuracy and sparseness in non-negative matrix factorization  \cite{foley}, etc. Furthermore, there are several variants of PSA in the literature. Implementation details of these variants can be found in \cite{article}.

\section{Mitigating Underestimation} \label{sec:mitigating_underestimation}

Given a dataset $D(X, Y)$, the goal of supervised ML is to learn an input-output mapping function $f: X \rightarrow Y$ that will generalize well on unseen data. From the optimization perspective, the learning process can be viewed as finding the best mapping that performs best in terms of how good the prediction model $f(X)$ does with regard to the expected outcome $Y$. For example, logistic regression, a well-studied algorithm for classification, can be formulated as an optimization problem where the objective is to learn the best mapping of feature vector $X$ to the target feature $Y$ through a probability distribution:
\begin{equation}
    P(Y = 1|X,\theta) = \frac{1}{1+e^{- \theta^T \cdot X}}
\end{equation}

where $\theta$ is obtained by solving an optimization problem to minimize log loss $L_{\theta}$:
\begin{equation}\label{eqn:logloss}
    L_{\theta} = -y \cdot log(P(Y = 1|X,\theta)) + (1-y) \cdot log(1-P(Y = 1|X,\theta))
\end{equation}

Solving this optimization problem directly without any explicit consideration of fairness could result in a model in which predictions show significant bias against a particular social group. To address this issue, one could directly modify the loss function in \ref{eqn:logloss} by adding a fairness constraint, see e.g. \cite{zafar2015}. The optimization is typically performed using variants of stochastic gradient descent (SGD). However, directly incorporating a fairness constraint into the loss function often results in a non-convex loss function. Consequently, optimizers that rely on the gradient of the loss function will struggle to converge due to the fact that non-convex functions have potentially many local minima (or maxima) and saddle points. Furthermore, many single objective optimization problems are NP-hard, and thus it is reasonable to expect that generating efficient solutions for MOOP is not easy \cite{https://doi.org/10.1002/(SICI)1099-1360(199801)7:1<34::AID-MCDA161>3.0.CO;2-6}. Due to this complexity, we propose a multi-objective optimization strategy using PSA to optimize for balance accuracy and underestimation. 

To illustrate the effectiveness of PSA to mitigate underestimation, we implement PSA for logistic regression. The aim is to find a set of $\theta$ that gives the highest balance accuracy and $US_{S} = 1$. Given a dataset $\mathcal{D}(X,Y,S)$, where $X$ represents the feature vector, target label $Y$ and sensitive attribute $S$, let $\hat{Y}$ be the prediction output of a logistic regression model $\mathcal{M}(\theta,X,S)$. The optimization problem can be formally defined as:

\begin{equation}\label{eqn:psa_obj}
    \theta = arg\max_\theta  \left(  BA(Y,\hat{Y}),\frac{1}{\left|1-US_{s}(Y,\hat{Y},S)\right|} \right)
\end{equation}

where $BA(Y,\hat{Y})$ represents the balance accuracy of classification prediction:
\begin{equation}\label{eqn:ba}
     BA(Y,\hat{Y}) = \frac{1}{2}\left( \frac{P[Y=1|\hat{Y}=1]}{P[Y=1]} + \frac{P[Y\ne1|\hat{Y}\ne1]}{P[Y\ne 1]}
     \right)
\end{equation}

The high-level description of our framework is presented in Algorithm \ref{algo}. The exact implementation can be found on \textsf{Github}\footnote{\url{https://github.com/williamblanzeisky/ParetoSimulatedAnnealing}}. Our algorithm is a modification of the PSA algorithm in \cite{https://doi.org/10.1002/(SICI)1099-1360(199801)7:1<34::AID-MCDA161>3.0.CO;2-6} tailored to optimize for balance accuracy and underestimation. It is worth noting that the choice of an appropriate cooling rate and initial temperature is crucial in order to ensure success of the algorithm \cite{article}. Our preliminary experiments suggest that there is a need for two temperature parameters $T_{ba}$ and $T_{us}$, due to the fact that balance accuracy is bounded in the range $[0,1]$ while underestimation is $[0,\infty]$. Thus, the acceptance probability defined in \ref{eqn:prob_accept} is modified to account for two initial temperatures:

\begin{equation}\label{eqn:prob_accept_modified}
    P(x,\bar{x},T,\lambda) = \acmISBN{}min \left\{ 1,exp\left(\sum_{i=0}^{m}\frac{ \lambda_{i} (f_{i}(x) - f_{i}(\bar{x}))}{T_{i}}\right) \right\}
\end{equation}

Given two initial temperatures $T_{ba}$ and $T_{us}$, a perturbation scale $\beta$ and a cooling rate $\alpha$, the PSA process consists of two loops (line 5 and 6); the first loop begins by randomly creating a set of solutions $Set_{\theta}$ with each solution representing a current solution $\theta$. Then, for each $\theta$ in $Set_{\theta}$, a candidate solution $\bar{\theta}$ is created by perturbing one of its dimensions with a  Gaussian noise $X  \sim \mathcal{N}(0,\beta^2)$. The candidate solution is accepted as the current best solution if it is "better" than the initial solution. If the candidate solution is dominated by the current best solution, it will only be accepted if $P(\theta,\bar{\theta},T_{ba},T_{us})$ is larger than a random value sampled from a uniform distribution $\mathcal{U}(0,1)$. Since the objective is to find the best solution $\theta$ that gives the best balance accuracy and underestimation, we set $\lambda$ to 1. This process is repeated for some iterations (see line 6). The second loop repeats this overall process (first loop)  and iteratively decrease the temperature by a cooling rate $\alpha$ until no further changes occur. The PSA algorithm will return a set of solutions $Set_{Pareto}$. We will then use the concept of Pareto optimality to select solutions that are non-dominated in terms of both criteria, referred to as the Pareto Set \cite{article}.  A solution is said to be Pareto Optimal if it is impossible to make one criterion better without making another criterion worse. 

\begin{algorithm}
\SetAlgoLined
\SetKwInOut{Input}{Input}
\SetKwInOut{Output}{Output}
\Input{A dataset $D (X,Y,S)$}
\Output{A Pareto front $Set_{Pareto}$}
 Initialize Pareto set $Set_{Pareto}$\;
 Randomly generate a set of initial solutions $Set_{\theta}$\;
 \For{each $\theta$ in $Set_{\theta}$}{
    Set initial temperatures $T_{ba}, T_{us}$, a perturbation scale $\beta$, a cooling rate $\alpha$ \;
    \While{$T_{ba}$ > 0 and $T_{us}$ > 0}{
    \For{some iterations}{
        Create a candidate solution $\bar{\theta}$\;
        \If{$\bar{\theta}$ improves on $\theta$}{
            Accept $\bar{\theta}$ as current best solution $\theta$\;
        \Else{
            Accept $\bar{\theta}$ with probability $P(\theta,\bar{\theta},T_{ba},T_{us})$\;}}
            
                                        }
        Decrease $T_{ba}$ and $T_{us}$ by $\beta$\;
                                        }
                                      }
        Filter non-dominated solutions from $Set_{Pareto}$
 
 \caption{High-level pseudocode for Pareto Simulated Annealing algorithm to mitigate underestimation.}
 \label{algo}
\end{algorithm}

\begin{figure*}[tp]
     \centering
     \includegraphics[width=\linewidth]{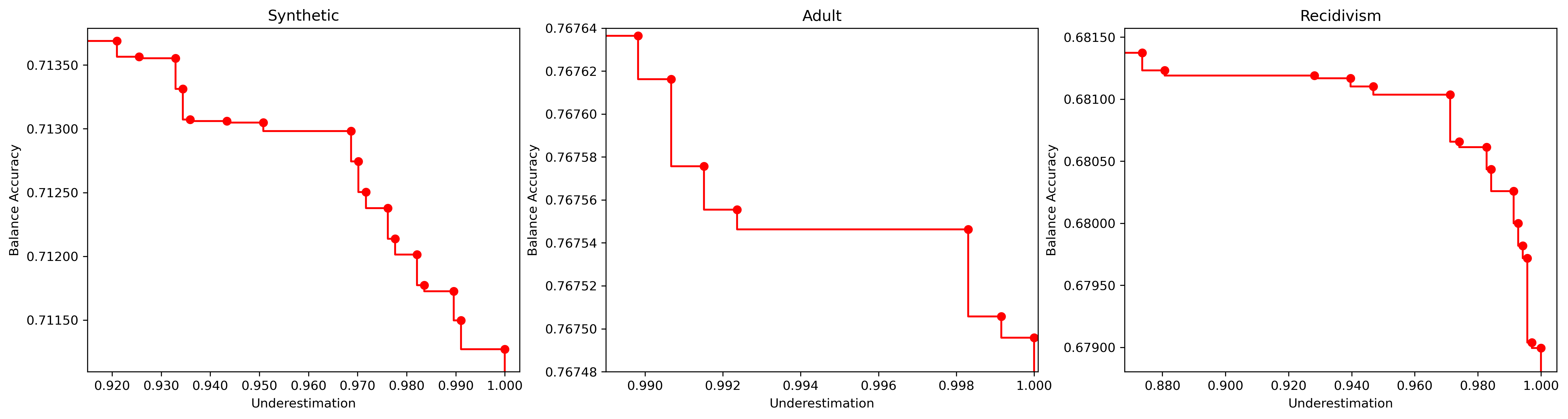}
     \caption{Pareto fronts for the three datasets.}
     \label{fig:pareto_front}
\end{figure*}


\begin{figure}[tp]
     \centering
     \includegraphics[width=0.9\linewidth]{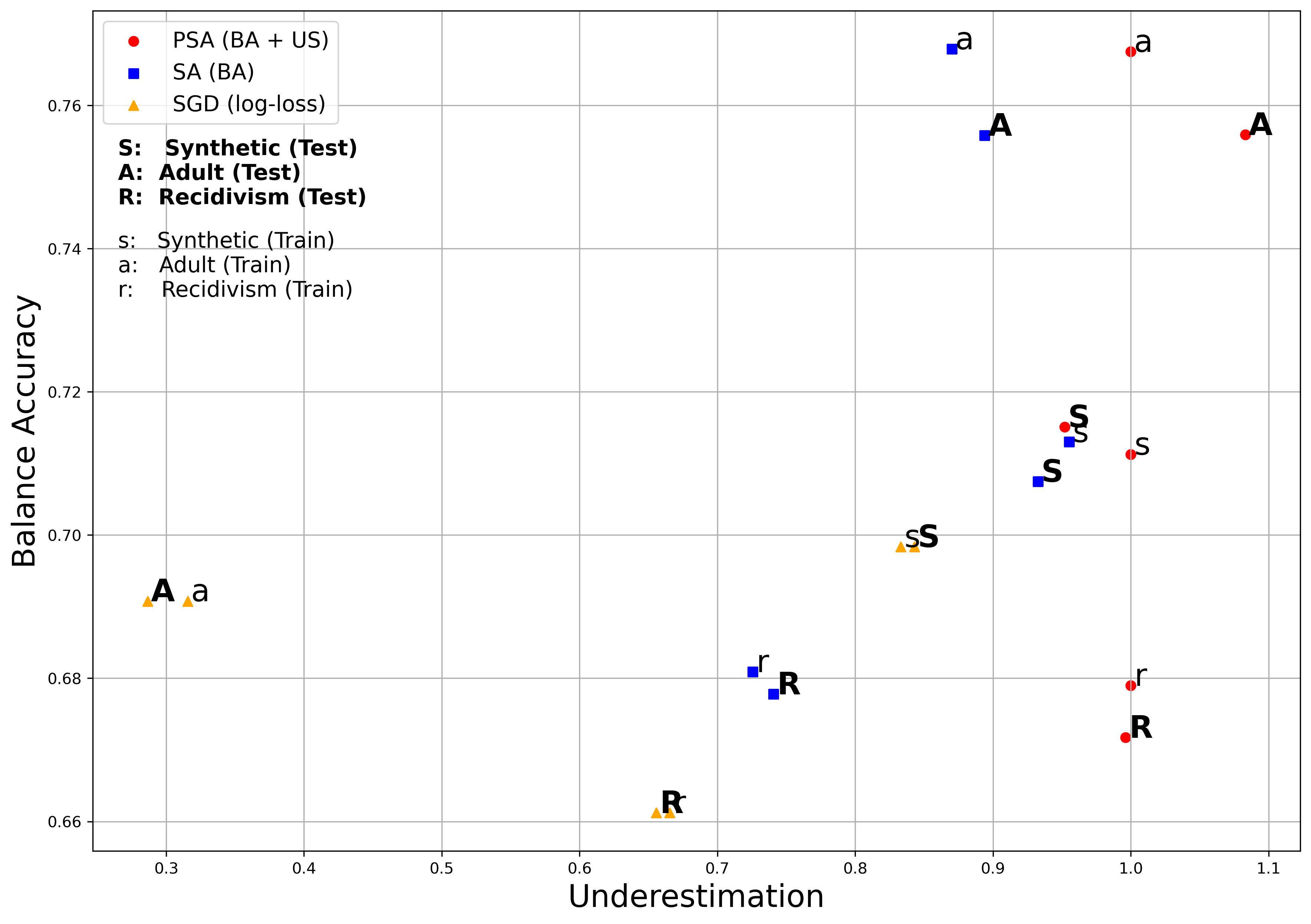}
     \caption{Overall results on the three datasets.}
     \label{fig:eval2}
\end{figure}

There are advantages and disadvantages of using PSA to mitigate underestimation. First, one could directly incorporate fairness constraints as the objective functions to be optimized without having to use a proxy convex function as an approximate measure of how good the predictions of a model is in terms of the expected outcome $Y$. Second, PSA is straight-forward to implement and its characteristic of accepting inferior solutions allows escaping from local minima/maxima, which is often desirable when the search space is large and the optimization functions are not convex. However, one limitation of PSA is that its implementation is problem-dependent, and the choice of appropriate parameters is itself a challenging task. Thus, preliminary experiments are often required \cite{article}.

\section{Experiments}\label{sec:experiments}

In this section, we experimentally validate our framework described in Section \ref{sec:mitigating_underestimation} on a synthetic dataset introduced in \cite{feldman2015certifying}, the Census Income dataset \cite{kohavi1996scaling} and a reduced version of the ProPublica Recidivism dataset \cite{dressel2018accuracy}. These datasets have been extensively studied in fairness research because there is clear evidence of negative legacy. Summary statistics for these datasets are provided in Table \ref{tab:datasets}.
For the Census Income dataset the prediction task is to determine whether a person earns more or less than \$50,000 per year based on their demographic information. We remove all the other categorical features in the Census Income (Adult) dataset except \textit{Sex}. The reduced and anonymized version of the Recidivism dataset includes seven features and the target variable is Recidivism. The goal of our experiment is to learn an underestimation-free classifier while maintaining a high balance accuracy score when \textit{Sex} and \textit{Caucasian} as sensitive feature $S$ for the Income and Recidivism datasets, respectively. 
To illustrate the effectiveness of the proposed strategy, we evaluate our framework \textit{PSA (BA+US)} with logistic regression optimized using SA on balance accuracy \textit{SA(BA)} and  \textsf{scikit-learn}\footnote{\url{https://scikit-learn.org/}} implementation of logistic regression optimized on log-loss \textit{SGD(log-loss)}.

\begin{table}
\begin{center}
\caption{Summary details of the three datasets.}\label{tab:datasets}
\begin{tabular}{ l | r | c| c}
Dataset & Samples & Features & \% Minority \\
\hline
Synthetic & 5000 & 3 & 50\% \\
(reduced) Adult - Income & 48,842 & 7 & 25\% \\
(reduced) Recidivism & 7,214 & 7& 45\% \\
\end{tabular}
\end{center}
\end{table}

For each of the datasets we use a 70:30 train test split. 
Figure \ref{fig:pareto_front} shows the Pareto front on the training sets. 
We can see from the Pareto fronts that the two criteria are in conflict.  
We can also see that one could achieve perfect underestimation ($US_{S} = 1$) without much loss in balance accuracy. 

Models with $US_{S} = 1$ on the training data were then tested on the test sets - see Figure \ref{fig:eval2}. 
For comparison we include results on models optimized using SGD (\textit{SGD(log-loss)}) and simulated annealing \textit{SA(BA)}. It is clear from Figure \ref{fig:eval2} that \textit{PSA(BA+US)} outperforms \textit{SA(BA)} and \textit{SGD(log-loss)} in all three datasets in terms of underestimation. In addition, Figure \ref{fig:eval2} also shows that a logistic regression classifier optimized using SA performs better than SGD. However, it is worth noting that SGD optimizes on log-loss while the simulated annealing directly optimizes on balance accuracy and underestimation, thus it makes sense that SA optimized algorithm performs better in terms of balance accuracy. To conclude, the evaluation shown in Figure \ref{fig:eval2} suggests that it is possible to achieve near-perfect underestimation score while maintaining good balance accuracy by explicitly considering underestimation bias in model training.

\section{Conclusion and Future work}

In this paper we present a multi-objective optimization strategy using Pareto Simulated Annealing that optimizes for both balanced accuracy \emph{and} underestimation. We demonstrate that our framework can achieve near-perfect underestimation while maintaining high balance accuracy on one synthetic and two real datasets. As an extension to this work, we plan to evaluate our strategy for other unsupervised ML algorithms, e.g., Neural Network, Decision Tree. In addition, we plan to evaluate our strategy on other datasets.

\begin{acks}
This work was funded by Science Foundation Ireland through the SFI Centre for Research Training in Machine Learning (Grant No. 18/CRT/6183) with support from Microsoft Ireland.
\end{acks}

\bibliographystyle{ACM-Reference-Format}
\bibliography{sample-base}

\end{document}